\documentclass{article}

\usepackage[preprint,nonatbib]{neurips_data_2024}

\usepackage[utf8]{inputenc} % allow utf-8 input
\usepackage[T1]{fontenc}    % use 8-bit T1 fonts
\usepackage{hyperref}       % hyperlinks
\usepackage{url}            % simple URL typesetting
\usepackage{booktabs}       % professional-quality tables
\usepackage{amsfonts}       % blackboard math symbols
\usepackage{nicefrac}       % compact symbols for 1/2, etc.
\usepackage{microtype}      % microtypography
\usepackage{xcolor}         % colors
\usepackage{graphicx}
\usepackage{booktabs}
\usepackage{enumitem}
\usepackage{parskip}
\usepackage{subcaption}
\usepackage{cleveref}
\usepackage{orcidlink}
\usepackage{tabularx}

\usepackage{multirow}

\title{Evaluating the Generation of Spatial Relations in Text and Image Generative Models}

\author{
 Sim Shang Hong \textsuperscript{1}\thanks{~~Equal contribution. Correspondence: clarence\_leesheng@mymail.sutd.edu.sg}
  \hspace{15pt} Clarence Lee \textsuperscript{1}\footnotemark[1] \hspace{15pt} Alvin Tan De Jun \textsuperscript{2}\footnotemark[1] \hspace{15pt} 
  ~Cheston Tan\textsuperscript{3}
     \\\\
     \hspace{-10pt} 
     \textsuperscript{1}Singapore University of Technology and Design  \\
     \hspace{-10pt} 
     \textsuperscript{2} Cornell University \\
     \hspace{-10pt} 
     \textsuperscript{3} Centre for Frontier AI Research, A*STAR 
}

\begin{document}
% \footnote[1]{Equal Contribution}
\maketitle
\begin{abstract}
\setcounter{footnote}{0}
Understanding spatial relations is a crucial cognitive ability for both humans and AI. While current research has predominantly focused on the benchmarking of text-to-image (T2I) models, we propose a more comprehensive evaluation~\footnote{\url{https://github.com/trolommonm/SpatialRelBench}} that includes \textit{both} T2I and Large Language Models (LLMs). As spatial relations are naturally understood in a visuo-spatial manner, we develop an approach to convert LLM outputs into an image, thereby allowing us to evaluate both T2I models and LLMs \textit{visually}. We examined the spatial relation understanding of 8 prominent generative models (3 T2I models and 5 LLMs) on a set of 10 common prepositions, as well as assess the feasibility of automatic evaluation methods. Surprisingly, we found that T2I models only achieve subpar performance despite their impressive general image-generation abilities. Even more surprisingly, our results show that LLMs are significantly more accurate than T2I models in generating spatial relations, despite being primarily trained on textual data. We examined reasons for model failures and highlight gaps that can be filled to enable more spatially faithful generations. 
\end{abstract}

\section{Introduction}
Understanding spatial relations, which emerges early in human development \cite{Hespos_Spelke_2004, rohlfing_nachtigäller_2016}, is critical for human cognition \cite{turan_mert}, and is required for a wide range of tasks such as robot navigation \cite{kuffner2005} and concept learning \cite{tenenbaum_2015}. Spatial relation understanding involves the usage of spatial prepositions -- words that establish relationships between nouns \cite{dictionary1989oxford} -- to articulate complex spatial dynamics between objects. These prepositions are not only crucial for human communication, but also serve to gauge the proficiency of generative models in understanding spatial relations. (``What I cannot create, I do not understand.'' -- Richard Feynman).

% \textbf{Success and shortcomings of T2I models}
Existing text-to-image (T2I) models have demonstrated impressive capabilities in rendering images from textual prompts. For example, T2I models are able to generate realistic images from complex prompts such as ``a cute corgi lives in a house made out of sushi''. Such impressive empirical results lead us to believe that T2I models are capable of understanding and reasoning about spatial relations. Despite these advancements, critical analyses of T2I models have uncovered inherent limitations. Insights from research focused on model development \cite{ramesh2022hierarchical, Betker_Goh_Jing} and various benchmarking studies \cite{gokhale2022benchmarking, conwell2022testing, huang2023t2i, cho2023dalleval, lee2023holistic, ghosh2023geneval,  saharia2022photorealistic} reveal that T2I models encounter difficulties with accurate prompt interpretation and spatial comprehension, hindering the ability to generate complex scenes precisely. 

% \textbf{Shortcoming of T2I benchmarking studies} 
While the above studies have been informative, analyses of spatial relations have been limited. Many studies which included spatial relations only cover a small number of such relations. Furthermore, many studies combine spatial relations with other factors, such as object counting and attribute binding, leading to confounds when assessing models' understanding of spatial relations. 

% \textbf{Shortcoming: LLM are under-explored} 
Although spatial relations in T2I models have been somewhat investigated, fewer papers have explored in detail spatial relations in LLMs. This is detrimental for a few reasons. Many T2I models rely on strong text encoders such as frozen LLMs for prompt understanding \cite{Betker_Goh_Jing, ai_2023}. Thus, the spatial relation understanding of T2I models will be affected by those LLMs. Previous research has also leveraged LLMs to enhance spatial understanding in diffusion models~\cite{lian2023llmgrounded, liu2022compositional}, highlighting interest in transferring knowledge from LLMs to improve image generation.

Better diagnosis of challenges faced by both T2I models and LLMs will enable us to disambiguate challenges which arise from either language understanding or the image generation process, allowing us to be more effective in leveraging LLMs to improve T2I performance. However, the inherent visual nature of spatial relations poses a challenge in evaluating text-based LLMs' understanding of them. Hence, we developed a method to convert LLM outputs into an image, allowing both T2I models and LLMs to be evaluated visually, as image generators.

We avoided the use of real-world objects, as this may invoke biases learnt during training. For example, ``dog under table'' occurs much more frequently than ``dog on table'' \cite{conwell2022testing}. Thus, if we prompt the model to generate ``dog on table'', we may just get ``dog under table'' simply due to data biases \cite{kamath2023s}. The use of simple visual scenes involving 3 basic geometric shapes and 10 prepositions allows more precise assessment of whether a model truly understands spatial relations.% By focusing on basic objects, we also enable consistent medium of evaluation between LLM and T2I.

Overall, our contributions are as follows: 
\begin{enumerate}[noitemsep,nolistsep,leftmargin=0.5cm]
    \item We developed the first visually-based investigation of spatial relation understanding, in LLMs, providing the full prompt set;
    \item We comprehensively benchmarked the generation of 10 important spatial relations in LLMs and T2I models;
    \item We pushed the boundaries of spatial relation understanding in generative models by systematically evaluating the composition of multiple spatial relations in the best performing generative models, thoroughly examining 90 distinct permutations.
\end{enumerate}

\begin{figure}
    \centering
    \includegraphics[width=\textwidth]{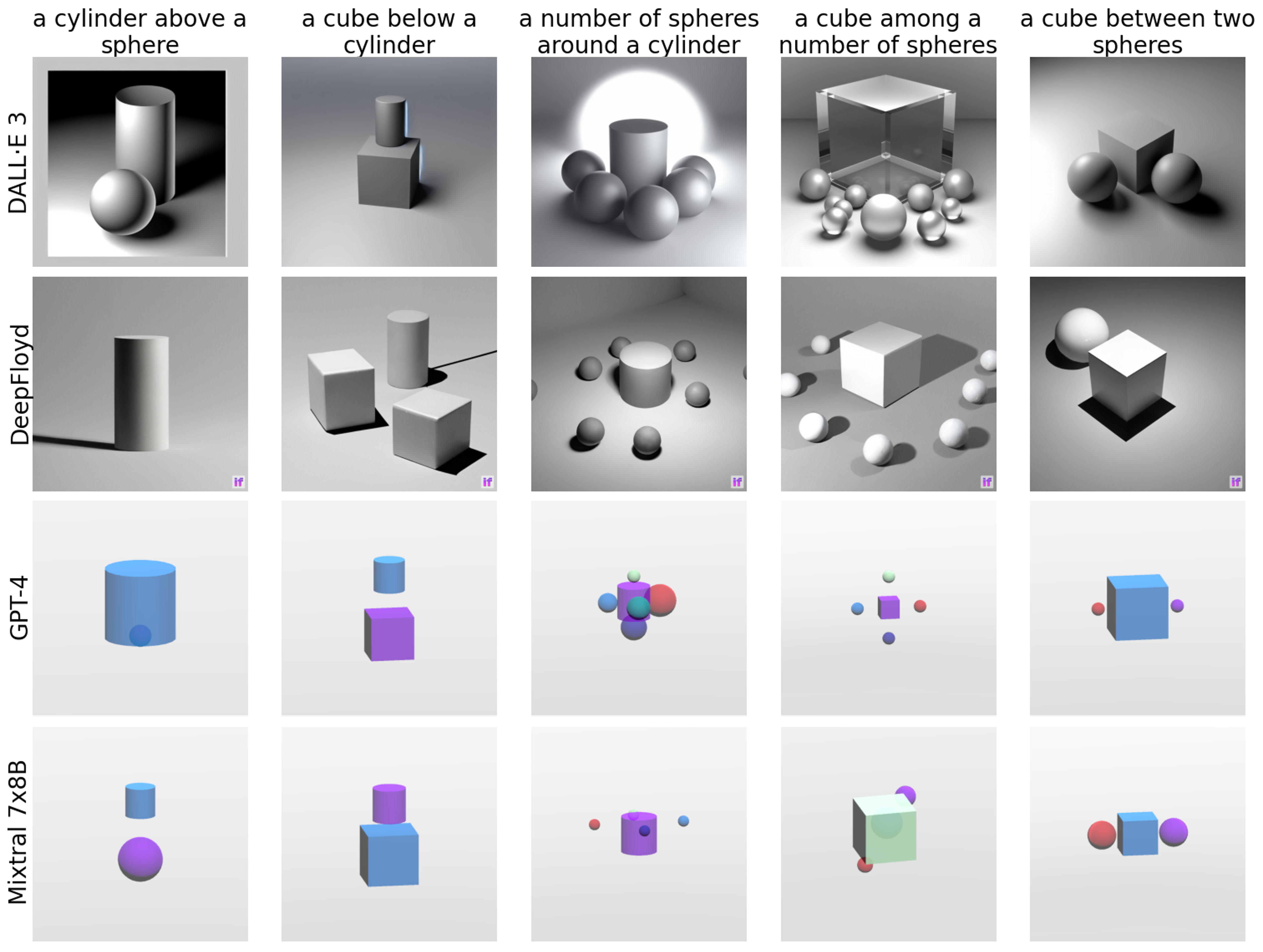}
    \caption{Images generated using both T2I and LLMs for some example prompts.}
    \label{fig:all_models_examples}
\end{figure}

\section{Related Work}

\subsection{Text-to-image models}
In this work, we compare three state-of-the-art text-to-image models frequently utilized in recent studies \cite{huang2023t2i, ghosh2023geneval, lee2023holistic}. \textbf{DALL·E 3} \cite{Betker_Goh_Jing} is a model designed to generate images from textual descriptions. It is known for its improved prompt-following abilities, achieved through training on highly descriptive synthetic captions derived from a bespoke image captioner, though its technical details remain undisclosed. \textbf{SDXL} (Stable Diffusion XL) \cite{podell2023sdxl} is a latent diffusion model that builds upon Stable Diffusion v2.1 with various architectural and training improvements. It comprises 3.4 billion parameters and is trained on Stability AI’s internal dataset. \textbf{DeepFloyd IF} \cite{ai_2023} is a cascaded pixel diffusion model capable of producing highly photorealistic images with a strong understanding of language. It consists of a frozen text encoder and three cascaded pixel diffusion modules totaling 6.5 billion parameters, trained on the LAION-A dataset containing 1 billion text-image pairs.

\subsection{Large Language Models}

We compare five state-of-the-art language models, including three closed-source models (GPT-4, GPT-3.5-turbo, and Gemini-pro) and two open-source models (Mixtral 8x7b and Llama2-chat-70b), chosen for their competitive performance in benchmarks such as MMLU \cite{hendrycks2020measuring} and GSM8k \cite{cobbe2021training}. \textbf{GPT-4} \cite{achiam2023gpt} is a multimodal model capable of processing both image and text inputs, with its technical details and parameter count undisclosed. \textbf{GPT-3.5-turbo} \cite{brown2020language} is an autoregressive model with 175 billion parameters, trained on a diverse range of NLP datasets including textual, question-answering, and cloze tasks. \textbf{Gemini-pro} \cite{team2023gemini} is a multimodal, multilingual model trained on data encompassing videos, images, audio, and text, also with undisclosed technical details and parameters. \textbf{Mixtral 8x7b} \cite{jiang2024mixtral} is a sparse mixture-of-experts model with 46.7 billion total parameters, utilizing 12.9 billion per token, and trained using supervised fine-tuning followed by Direct Preference Optimization\cite{rafailov2024direct}. \textbf{Llama2-chat-70b} \cite{touvron2023llama}, part of the Llama 2 series, is a 70 billion parameter model fine-tuned for dialogue with purely textual data, making it the largest model in the series.

\subsection{Existing spatial understanding benchmarks}
Current methods for evaluating spatial understanding in text-to-image models follow three main approaches. The first employs object detectors to assess the accuracy of spatial relations in generated images \cite{huang2023t2i, minderer2022simple, gokhale2022benchmarking, ghosh2023geneval, cho2023dalleval}. The second measures image-text similarity through CLIPScore \cite{hessel2022clipscore}, evaluating the similarity between image and text embeddings, or through image captioning-based evaluation \cite{hong2018inferring}, using metrics like BLEU \cite{papineni-etal-2002-bleu}, METEOR \cite{banerjee-lavie-2005-meteor}, and ROUGE \cite{lin-2004-rouge}. The third approach utilizes a multimodal LLM, such as Mini-GPT-4, to automatically evaluate spatial relations \cite{huang2023t2i, lu2024llmscore}, where the model describes the image and predicts the image-text alignment score.

Current methods to evaluate spatial reasoning in LLMs follow two main approaches: (1) evaluating spatial reasoning abilities based on its ability to accurately infer physical and abstract spatial relations \cite{comsa-narayanan-2023-benchmark}, and (2) evaluating the LLM's ability to navigate a toy grid puzzle described in text \cite{anonymous2023evaluating}. 

\subsection{Visual grounding for text-to-image models}

In order to address challenges in generating spatially faithful images, \textbf{L}L\textbf{M}-grounded \textbf{D}iffusion (LMD) \cite{lian2023llmgrounded} was proposed to guide the diffusion model through an LLM generated scene layout comprising of captioned bounding boxes generated from the given prompt. LayoutGPT \cite{feng2023layoutgpt} was another method proposed to guide text-to-image generation by providing layouts in form of CSS generated from given prompts. Visual ChatGPT \cite{wu2023visual} was proposed to allow users to iteratively interact with a suite of Visual Foundaional Models (VFMs) through the conversational nature of ChatGPT. \textbf{G}enerating \textbf{I}mages with \textbf{L}arge \textbf{L}anguage Models (GILL) \cite{koh2023generating} was proposed to fuse LLMs with text-to-image models such as Stable Diffusion to achieve strong performance on text-to-image generation tasks. Another viable approach is via locally conditioned diffusion \cite{Po2023Compositional3S} where the diffusion model is able to take in 3D bounding boxes and text prompts to generate coherent 3D scenes and objects. 

\subsection{Evaluation of text-to-image models}
Research focused on model development \cite{ramesh2022hierarchical, Betker_Goh_Jing} has highlighted that T2I models struggle with accurate prompt interpretation and spatial comprehension. To more comprehensively evaluate the prompt following abilities of T2I models, benchmarks such as Drawbench \cite{saharia2022photorealistic}, T2I-CompBench \cite{huang2023t2i} and HEIM \cite{lee2023holistic} have been released. To further examine the spatial understanding of T2I models, PaintSkills \cite{cho2023dalleval} has been proposed to evaluate T2I models' visual reasoning capabilities. \textit{Conwell at el.} \cite{conwell2022testing} studied relational understanding of DALL·E 2 using a set of 15 basic relations. 

\section{Methods}

\subsection{Setup} \label{sec:setup}
In our experiments, we used both human raters and GPT-4V to assess if the spatial relations were accurately represented in the images generated by the 8 generative models. 

Although there are over 100 prepositions in total \cite{TPP}, we selected the 10 most common ones from the literature \cite{Retz-Schmidt_1988, coventry1994spatial, Zwarts_2017, agrawal2023stupd} to assess the models' spatial understanding. These prepositions are: \textit{above, below, on, among, around, beside, in front of, behind, between, inside}. We specifically chose prepositions that can be paired, such as above/below, among/around, and in front of/behind.

% While there are more than 100 prepositions in total \cite{TPP}, we curated the 10 most common prepositions based on existing literature \cite{Retz-Schmidt_1988, coventry1994spatial, Zwarts_2017, agrawal2023stupd} to gauge the models' spatial understanding abilities. The 10 spatial prepositions are: \textit{above, below, on, among, around, beside, in front of, behind, between, inside}. We chose certain prepositions that could be paired together such as above/below, among/around and in front of/behind.

We also chose to use three basic geometric objects: \textit{sphere, cube, cylinder} to minimize the influence of object complexity on the analysis of the spatial relation. By using these basic shapes, we were also able to seamlessly integrate them with LLMs as the three objects exist as primitives in Unity, allowing us to ensure a consistent evaluation method for both T2I and LLMs.

% The use of simple geometric objects helps to create a common medium to measure both LLMs and T2I models \textcolor{magenta}{(CT: because LLMs can only generate 3D locations, but not 3D objects?)}, while isolating the models' spatial understanding ability by abstracting all details away. \textcolor{magenta}{(CT: details of what?)}

\subsection{Text prompts} \label{sec:prompt_setup}
We evaluated the models using two types of text prompts as follows:

% corresponding to single spatial relations and multiple spatial relations.

\textbf{Single spatial relations:} In our basic setup, we create prompts containing a single relation, termed as \textit{simple prompts}, which have the following structure: \textit{``\{\textcolor{red}{object1}\} \{\textcolor{blue}{spatial preposition}\} \{\textcolor{magenta}{object2}\}''}. An example would be ``a \textcolor{red}{cube} \textcolor{blue}{in front of} a \textcolor{magenta}{sphere}''. In total, eight images were generated for each spatial relation. To attain the images, four distinct prompts were attained by choosing randomly  four of the $^3P_2=6$ permutations for \textcolor{red}{object1} and \textcolor{magenta}{object2}. We then prompt the models to generate two variations of the scene for each of the four prompts, using a temperature of 1.0 for the LLMs to achieve diversity, and enforce different generations by iteratively generating until two distinct results are obtained. This results in 80 images per model in total (8 images x 10 prepositions) for simple prompts.

% \textbf{Single spatial relations:} For each spatial preposition, we generated distinct prompts through permuting the three different objects provided, \textcolor{magenta}{(CT: shouldn't there be 6 prompts? Explain.)}. The resulting prompts, hereby referred to as \textit{simple prompts}, have the following structure: \textit{``\{\textcolor{red}{object1}\} \{\textcolor{blue}{spatial preposition}\} \{\textcolor{magenta}{object2}\}''}. An example would be ``a \textcolor{red}{cube} \textcolor{blue}{in front of} a \textcolor{magenta}{sphere}''. In total, each spatial relation generates 8 images each, which is created by generating a set of four prompts twice through the use of a temperature value of 1.0 Each prompt was given to every model twice, and we used a temperature value of 1.0 to get different generations. Hence, for every preposition, we have 8 images. This results in 80 images in total (8 images x 10 prepositions) for simple prompts.

\textbf{Multiple spatial relations:} To stress-test the models, we evaluated the top two performing generative models using \textit{complex prompts}. A \textit{complex prompt} was defined as comprising two spatial prepositions and three objects in the scene, and is represented by the template: \textit{``\{\textcolor{red}{object1}\} \{\textcolor{blue}{spatial preposition 1}\} \{\textcolor{magenta}{object 2}\}, with \{\textcolor{orange}{object 3}\} \{\textcolor{green}{spatial preposition 2}\} \{\textcolor{red}{object 1}\}''}. An example is ``a \textcolor{red}{cube} \textcolor{blue}{above} a \textcolor{magenta}{sphere}, with a \textcolor{orange}{cylinder} \textcolor{green}{beside} the \textcolor{red}{cube}''. In total, there are $^{10}P_{2} = 90$ possible permutations of 10 spatial prepositions with $^3P_3 = 6$ possible arrangements of three objects in a sentence creating 540 possible unique prompts. We do not repeat the same relation in a prompt as we want to test the ability to compose different relations together. For each prompt, we generated 4 different images by permuting the order of objects, leading to 360 (90 permutations of prompts x 4) images for complex prompts.

% \begin{table}[h!]
%     \centering
%     \newcolumntype{Y}{>{\centering\arraybackslash}X}
%     \caption{Examples of complex prompts with GPT-4 and Mixtral 8x7b, randomly sampled}
%     \begin{tabularx}{\textwidth}{YYYY}
%     \centering GPT-4 & GPT-4 & Mixtral 8x7b & Mixtral 8x7b\\
%     \includegraphics[width=0.23\textwidth]{figures/examples/complex_prompt/gpt-4/1_a cube above a sphere, a cylinder beside the cube.png}& \includegraphics[width=0.23\textwidth]{figures/examples/complex_prompt/gpt-4/1_a number of cubes around a sphere, with a cylinder below the sphere.png} & \includegraphics[width=0.23\textwidth]{figures/examples/complex_prompt/mixtral/1_a cylinder below a cube, with the cylinder being between two spheres.png} & \includegraphics[width=0.23\textwidth]{figures/examples/complex_prompt/mixtral/0_a cube in front of a cylinder, with the cube among a number of spheres.png} \\
%     ``A cube above a sphere, with a cylinder beside the cube'' & ``A number of cubes around a sphere, with a cylinder below the sphere'' & ``a cylinder below a cube, with the cylinder being between two spheres'' & ``a cube in front of a cylinder, with the cube among a number of spheres'' \\
%     \end{tabularx} \\
%     \label{table:example_complex_prompts}
% \end{table}

\subsection{Pipeline}

\begin{figure}[t]
    \centering
    \includegraphics[width=0.8\textwidth]{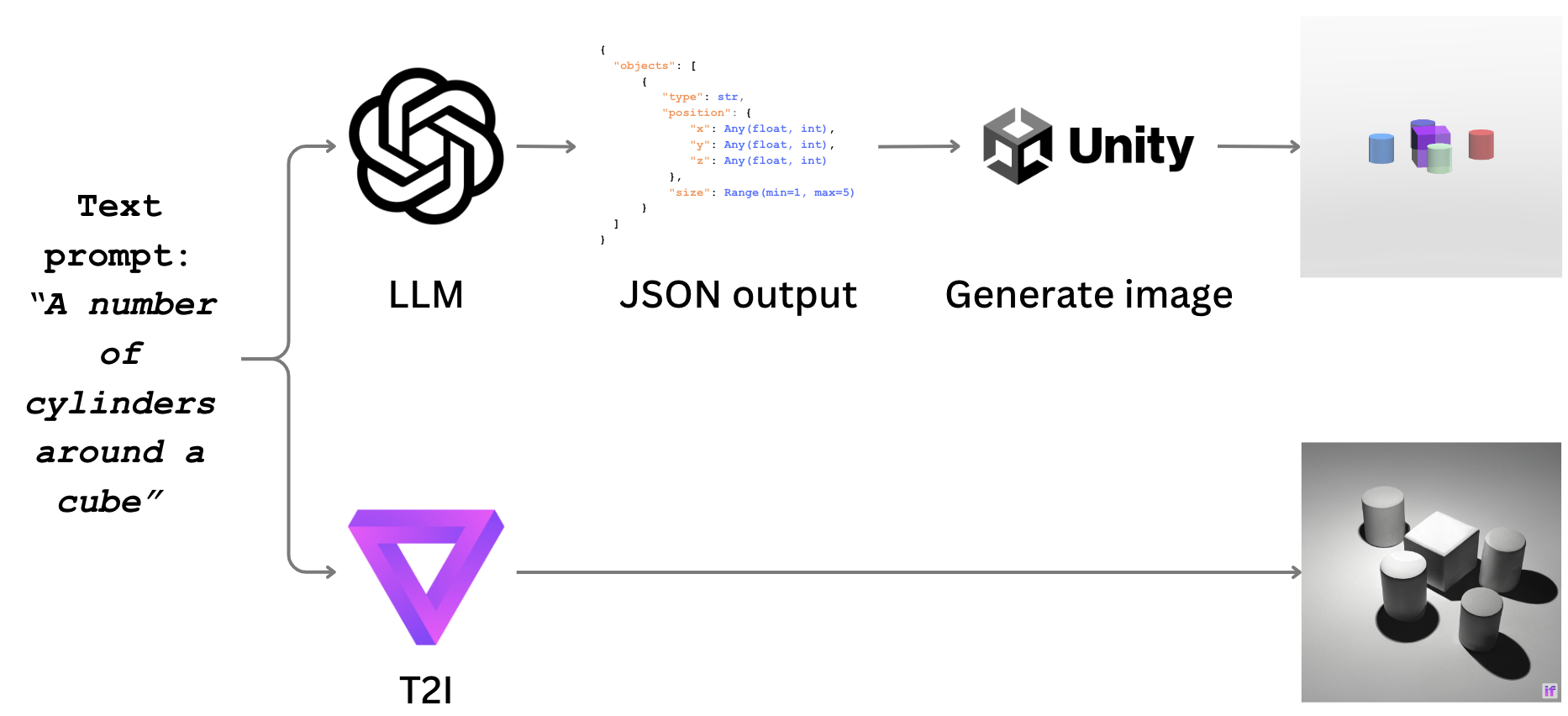}
    \caption{Overall image-generation pipelines for LLMs and T2I models.}
    \label{fig:pipeline}
\end{figure}

We input text prompts to the LLMs, instructing them to use the left-hand coordinate system for standardizing viewing perspective. For example, ``a cube in front of a sphere''. This prompt can have two distinct interpretations: one from the viewer's perspective, where the cube appears closer to the viewer and the sphere further away; and another from the sphere's perspective, implying the cube is positioned closer to the sphere but potentially seen as behind the sphere from the viewer's viewpoint. Standardizing the perspective minimizes ambiguity, ensuring any inaccuracies in generated relations are due to model misunderstandings rather than viewpoint inconsistencies. The generated JSON defining the 3D scene is rendered in Unity (see \Cref{fig:pipeline}), to ensure fair comparison with the T2I models, which cannot generate json. Similarly, each T2I model receives a text prompt with instructions to standardize perspective to that of the viewer. The prompts can be found in Supplementary Materials.

% \begin{figure}[t]
%     \centering
%     \includegraphics[width=0.9\textwidth]{figures/left-hand-coordinate-system.png}
%     \caption{Standardised left-hand coordinate system to reduce ambiguity in interpreting spatial relations.}
%     \label{fig:coordinate_system}
% \end{figure}

\subsection{Human Evaluation} \label{sec:human_eval}
For evaluation, anonymous Amazon Mechanical Turk workers rated the images using the interface shown in the Supplementary Materials. To the best of our knowledge, there are no significant ethical impact as a result of our study and no personal data was collected for the study. In order to verify rating quality, we measured the inter-rater agreement, using Fleiss' kappa $\kappa$ \cite{gisev2013interrater}. Fleiss' kappa is a well-known statistical measure for assessing reliability of agreement between a fixed number of three or more raters when using a nominal scale. 

\textbf{Single spatial relations:} Raters were presented with 80 images to rate per batch. To ensure evaluation accuracy, control examples (with unambiguous ground-truth ratings) comprising 10\% of the total examples were included. Only batches that had $\geq 87.5\%$ (i.e. 7 out of 8) accurate ratings for the control examples were accepted. Each image was rated by 10 raters.

\begin{figure}[htb!]
     % \centering
     \begin{subfigure}[t]{0.23\linewidth}
         \centering
         \includegraphics[width=\textwidth]{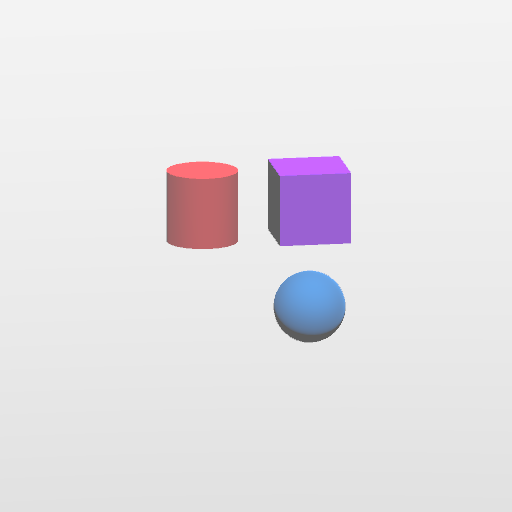}
         \caption{\textbf{GPT-4.} A cube above a sphere, with a cylinder beside the cube}
     \end{subfigure}
     \hfill
     \begin{subfigure}[t]{0.23\linewidth}
         \centering
         \includegraphics[width=\textwidth]{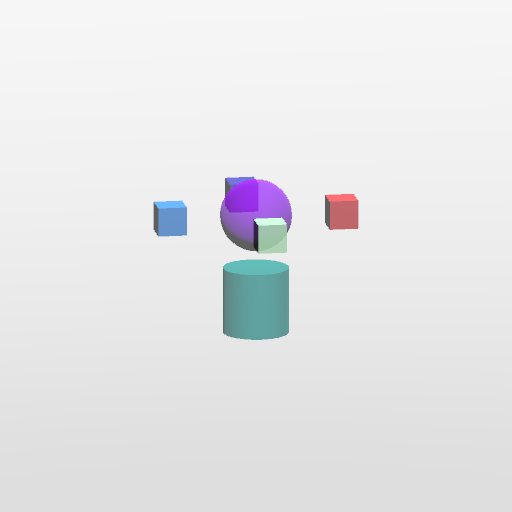}
         \caption{\textbf{GPT-4.} A number of cubes around a sphere, with a cylinder below the sphere}
     \end{subfigure}
     \hfill
     \begin{subfigure}[t]{0.23\linewidth}
         \centering
         \includegraphics[width=\textwidth]{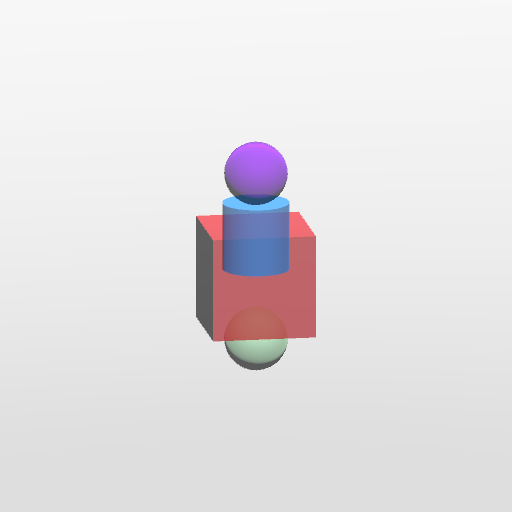}
         \caption{\textbf{Mixtral.} A cylinder below a cube, with the cylinder being between two spheres}
     \end{subfigure}
     \hfill
     \begin{subfigure}[t]{0.23\linewidth}
         \centering
         \includegraphics[width=\textwidth]{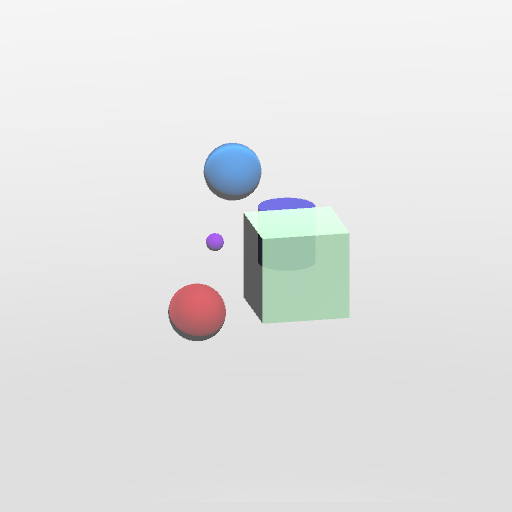} 
         \caption{\textbf{Mixtral.} A cube in front of a cylinder, with the cube among a number of spheres}
     \end{subfigure}
        \caption{Examples of images generated using GPT-4 and Mixtral 8x7b for complex prompts.}
        \label{table:example_complex_prompts}
\end{figure}

\textbf{Multiple spatial relations:} For the complex prompt setup, raters rated 90 images per batch, and control examples comprising 10\% (9 out of 90) of the total size of the batch were included. Only batches with $\geq 88.9\%$ (i.e. 8 out of 9) accuracy on control examples were used. \Cref{table:example_complex_prompts} shows some examples of complex prompts.

\subsection{Automatic evaluation using GPT-4V}

% Given the prowess of GPT-4 in generating spatial relationships (as we will see in Section \ref{sec:overall results}), we were motivated to test how well GPT-4V may perform in rating images. We draw upon previous work\cite{huang2024t2i} to assess the capabilities of current automatic evaluation tools. To evaluate each image, we passed the prompts as detailed in \Cref{table:automatic_eval} and images to GPT-4V. We only performed automatic evaluation on simple prompts. Two evaluation setups were used to gauge the model's ability to rate images. The first setup involved using GPT-4V to generate a caption for the image (using Prompt 1). The image, rating scale and generated caption are then passed as inputs to GPT-4V to derive a rating score (using Prompt 2). The second setup omitted the caption generation step, and provided only the image and rating scale. 

Given GPT-4's ability to generate spatial relationships (see Section \ref{sec:overall results}), we were motivated to evaluate GPT-4V's performance in rating images. Drawing upon previous work \cite{huang2023t2i}, we assessed current automatic evaluation tools by providing prompts (as detailed in \Cref{table:automatic_eval}) and images to GPT-4V. We focused on simple prompts and utilized two evaluation setups. The first involved GPT-4V generating a caption for the image (using Prompt 1). The image, rating scale and generated caption are then passed as inputs to GPT-4V to derive a rating score (using Prompt 2). The second setup omitted the caption generation step, and provided only the image and rating scale. 

% \begin{table}[hbt!]
\begin{table}[t]
  \caption{Automatic evaluation scale for simple prompts. Setup with captioning: prompt 1 + prompt 2. Setup without captioning: prompt 2 only.}
  \label{table:automatic_eval}
  \centering
  \fontsize{8}{10}\selectfont
  \begin{tabular}{|p{0.99\linewidth}|}
    \hline
    \textbf{Prompt 1} = ``You are my assistant to identify 3D objects and their spatial layout in the image. Briefly describe the image, focusing on the spatial position of each 3D object, within 50 words.''\\
    \hline
    \textbf{Prompt 2} = ``According to the image and your previous answer, evaluate if the text \{text\} is correctly portrayed in the image. \\
    Give a rating (A, B, C, D), according to the criteria:\\
    A: Spatial relationship is correct, and the number and type of objects is correct\\
    B: Spatial relationship is correct, but one type of object is wrong, or number of objects is wrong\\
    C: Spatial relationship is wrong, but the type and number of objects are correct\\
    D: Spatial relationship is wrong, and the type of objects are wrong\\
    Provide your analysis and explanation in JSON format with the following keys: rating (e.g., C), explanation (within 20 words).''\\
    \hline
  \end{tabular}
\end{table}

To see how well generative models perform as an evaluator, we measured Cohen's kappa between GPT-4's automatic scores and human ratings. We chose Cohen's kappa over Fleiss' kappa due to the two sets of nominal ratings. We obtained a single rating (A-D) for each image and model by taking the mode of 10 human raters. We then calculated the agreement between the 80 average human ratings and 80 GPT-4V ratings. To further assess GPT-4V as a human evaluation substitute, we treated GPT-4V as the 11th evaluator in a leave-one-out process. This method compares each rater against the mode of the remaining 10 using Cohen's kappa, providing a thorough examination of GPT-4V's consistency with human ratings. Additional details are attached in the Supplementary Materials.

We decided not to use the similarity in latent representations of the text and image, in a CLIP \cite{radford2021learning} based fashion, to do the evaluation. This is because we found that CLIP's evaluations do not correlate well with the rankings of the human annotators. Previous work \cite{lin2024evaluating} also show the tendency for CLIP to operate as a bag of words, conflating prompts such as ``a cube above a sphere'' and ``a sphere above a cube'', which hinders its ability to accurately recognize and evaluate the organization of objects in the context of spatial relations.

\section{Results: Single spatial relationship}
We evaluated 8 generative models on the set of 80 simple prompts and discuss key findings in \Cref{fig:simple_results} and \Cref{tab:percentage_A_simple}. On average, the Fleiss' kappa score was $\kappa = 0.52$, which indicates a moderate level of agreement among the 10 raters.
%based on uniform weights on the different classes. Fleiss' kappa is

\subsection{Overall results}\label{sec:overall results}

\begin{figure}[ht!]
    \centering
    \includegraphics[width=\textwidth]{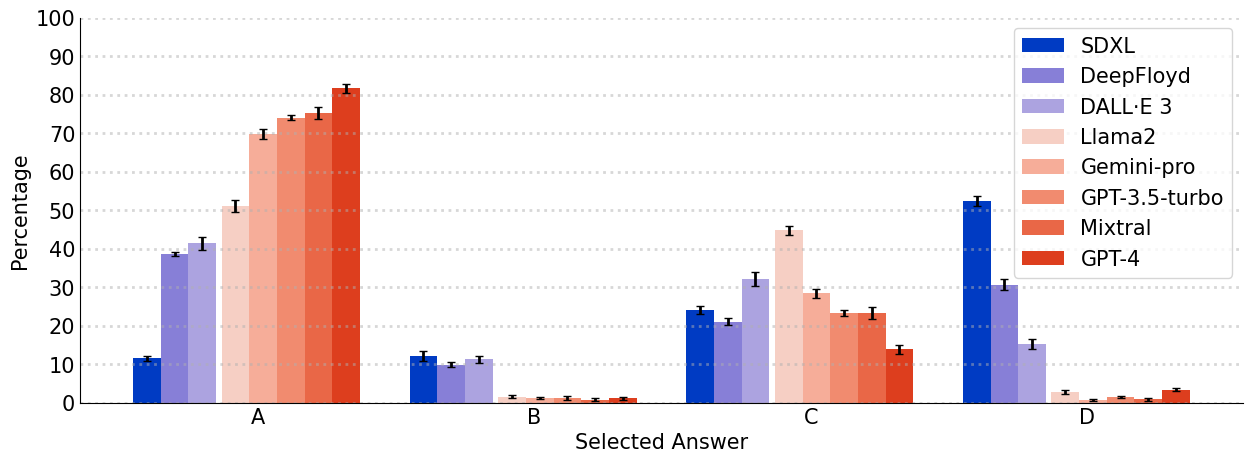}
    % \caption{Percentage of ratings per model for each answer}
    \caption{\textbf{Distribution of human ratings of images generated for simple prompts.} Detailed descriptions of the four ratings are as follows. \textbf{A:} Spatial relationship is correct, and the number and type of objects is correct. \textbf{B:} Spatial relationship is correct, but one type of object is wrong, or number of objects is wrong. \textbf{C:} Spatial relationship is wrong, but the type and number of objects are correct. \textbf{D:} Spatial relationship is wrong, and the type of objects are wrong.}
    \label{fig:simple_results}
\end{figure}

\textbf{GPT-4 is the most capable model, followed by Mixtral.} In \Cref{fig:simple_results}, GPT-4 achieved the highest accuracy at 82.6\%, followed by Mixtral at 79.0\% accuracy. In \Cref{tab:percentage_A_simple}, GPT-4 topped 6 out of 10 spatial prepositions, followed by Mixtral, which topped 2 out of 10 spatial prepositions. While Mixtral is significantly smaller than the other LLMs, and only uses 12.9B parameters during inference, it outperformed all other models except GPT-4 (which was only 3.6\% better).

\begin{table}[ht!]
  \caption{Percentage of ``A'' ratings for simple prompts across each model and preposition. \textbf{Bold} indicates maximum for each preposition in each category (T2I or LLM). \underline{Underlining} indicates overall maximum for each preposition. \textcolor{blue}{Blue} indicates notable maximum in T2I average. \textcolor{red}{Red} indicates notable minimum in LLM average.}
  \label{tab:percentage_A_simple}
  \centering
  \fontsize{7.8}{9.5}\selectfont
  \begin{tabularx}{\textwidth}{lcccp{0.9cm}<{\centering}cccccp{0.9cm}<{\centering}}
\toprule

& \multicolumn{4}{c}{T2I} & \multicolumn{6}{c}{LLM} \\
\cmidrule(lr){2-5}
\cmidrule(lr){6-11}
Relation & SDXL & DF & DALL·E 3 & \textbf{Avg} & Llama2 & Gemini & GPT-3.5 & Mixtral & GPT-4 & \textbf{Avg} \\ 
\midrule
Above & 11.2 & 36.2 & \textbf{43.8} & 30.4 & 63.7 & 86.2 & 83.8 & \textbf{\underline{95.0}} & 93.8 & 84.5\\ 
Below & 7.5 & 16.2 & \textbf{23.8} & 15.8 & 33.8 & 66.2 & 65.0 & 90.0 & \textbf{\underline{91.2}} & 69.2\\
On & 25.0 & \underline{\textbf{72.5}} & 17.5 & 38.3 & 48.8 & 53.8 & 45.0 & 48.8 & \textbf{60.0} & \textcolor{red}{51.3}\\ 
Among & 25.0 & \textbf{83.8} & 78.8 & \textcolor{blue}{62.5} & 66.2 & 82.5 & 72.5 & \textbf{\underline{87.5}} & 63.7 & 74.5\\ 
Around & 13.8 & 77.5 & \textbf{90.0} & \textcolor{blue}{60.4} & 66.2 & 81.2 & 92.5 & 91.2 & \textbf{\underline{95.0}} & 85.2\\ 
Beside & 42.5 & 56.2 & \textbf{81.2} & \textcolor{blue}{60.0} & 52.5 & 80.0 & 91.2 & 80.0 & \textbf{\underline{96.2}} & 80.0\\ 
In Front Of & 16.2 & \textbf{50.0} & 41.2 & 35.8 & 56.2 & 72.5 & \underline{\textbf{81.2}} & 71.2 & 67.5 & 69.7\\ 
Behind & 17.5 & 28.7 & \textbf{36.2} & 27.5 & 63.7 & 60.0 & 65.0 & 62.5 & \underline{\textbf{72.5}} & 64.7\\ 
Between & 18.8 & 13.8 & \textbf{46.2} & 26.3 & 61.3 & 68.8  & 88.8 & 81.2 &  \underline{\textbf{97.5}} & 79.5\\ 
Inside & 35.0 & 15.0 & \textbf{50.0} & 33.3 & 67.5 & 80.0 & 80.0 & 82.5 & \underline{\textbf{88.8}} & 79.8\\ 
\midrule
\textbf{Avg} & 21.2 & 45.0 & \textbf{50.9} & 39.0 & 58.0 & 73.1 & 76.5 & 79.0 & \textbf{\underline{82.6}} & 73.8\\
\bottomrule
  \vspace{-0.5pt}
  \end{tabularx}
  \footnotesize{DF = DeepFloyd,  GPT-3.5 = GPT-3.5-turbo, Gemini = Gemini-pro, Llama2 = Llama2-70b-chat, Mixtral = Mixtral-8x7B}
\end{table}

\subsection{Comparison of LLM and T2I results} \label{sec:comparison_t2i_llm_results}
\textbf{LLMs generate single spatial relations more accurately than T2I models.} In \Cref{fig:simple_results}, the percentage of correct answers by LLMs (red bars under option A) is significantly higher than the percentage of correct answers by T2I models (blue bars under option A). We observe a similar trend in \Cref{tab:percentage_A_simple}, with DALL·E 3, the best T2I model, performing worse than than Llama-70b-chat, the worst LLM.

\textbf{LLMs are less prone to make errors regarding the type and number of objects as compared to T2I.} In \Cref{fig:simple_results}, we observe that LLMs have less B and D ratings, which requires the model to make an error in the number and type of objects, as compared to T2I. This indicates that T2I models may struggle with prompt following for both spatial relations and the type and number of objects, while LLMs mainly struggle with representing the spatial relations. 

\textbf{The accuracy for \textit{Below} is significantly lower than \textit{Above} for both T2I and LLMs.} In \Cref{tab:percentage_A_simple}, \textit{Below} achieved 69.2\% and 15.8\% while \textit{Above} achieved 84.5\% and 30.4\%, for LLM and T2I respectively. We hypothesize that the disparity is due to the higher frequency of \textit{Above} in training data. \textit{Kamath et al.} reported \textit{Above} appears twice as often as \textit{Below} in the LAION-2B dataset \cite{kamath2023s}, potentially contributing to the models' poorer performance for \textit{Below}.

% A study done by \textit{Kamath at el.} reported that \textit{Above} occurs twice as frequently as \textit{Below} in the LAION-2B dataset \cite{kamath2023s}. This bias in training data could contribute to the models' poorer performance for \textit{Below}. 

\textbf{LLMs and T2I models tend to struggle with \textit{In Front Of} and \textit{Behind}.} In \Cref{tab:percentage_A_simple}, \textit{In Front Of} and \textit{Behind} achieved 69.7\% and 64.7\% accuracy respectively for LLMs. This is on average 7\% lower than the LLM average of 73.8\%. Similarly, \textit{In Front Of} and \textit{Behind} achieved 35.8\% and 27.5\% accuracy respectively for T2I models, which is on average 7\% lower than the T2I model average of 39.0\%.  \textit{Kamath at el.} attribute this to the ambiguity of prepositions and inconsistencies in dataset annotation, leading to poor model performance on viewer-perspective references \cite{kamath2023s}.

% Despite controlling the perspective during generation and providing instructions to raters to take the perspective of the viewer, there is still inherent ambiguity in how raters interpret the perspective of the image. As such, the ambiguity in annotator interpretation could have led to the lower performance in ``in front of'' and ``behind''

% \textbf{LLM tend to struggle with In Front Of and Behind.} We hypothesize that this is due to the model being confused about perspective. For example, "A cube in front of a sphere". There are two ways of interpreting this, one from the viewer's point of view where the cube will be in the foreground and the sphere in the background. The other interpretation is where the cube is in front of the sphere, from the sphere's perspective. From the viewer's point of view, it can appear as the cube is behind the sphere instead. \\

% We hypothesize that this arises from ambiguity in perspective, inherent to the prepositions. For example, ``a cube in front of a sphere''. This prompt can have two distinct interpretations: one from the viewer's perspective, where the cube appears closer to the viewer and the sphere further away; and another from the sphere's perspective, implying the cube is positioned closer to the sphere but potentially seen as behind the sphere from the viewer's viewpoint. This dual interpretation suggests that LLMs' confusion may stem from ambiguity in determining the correct perspective. 

% \subsection{LLM results}\label{sec:comparison_t2i_llm_results}

\textbf{\textit{On} is challenging for LLMs.} In \Cref{tab:percentage_A_simple}, \textit{On} achieves an average accuracy of 51.3\%, 22.5\% lower than the LLM average of 73.8\%. \textit{On} is a preposition that requires objects to be in contact, without any overlap. This requires precision in object placements, a task that is difficult for LLMs.
% due to the amount of precision needed to place the objects in contact with each other without overlap. The precision makes it a difficult task for LLMs.  

% \textbf{LLMs can follow instructions fairly well.} It makes minimal errors with regards to generating jsons.

% \subsection{T2I results}

\textbf{\textit{Among}, \textit{Around} and \textit{Beside} are the best performing spatial prepositions for T2I models.} In \Cref{tab:percentage_A_simple}, \textit{Among}, \textit{Around} and \textit{Beside} achieved 62.5\%, 60.4\% and 60.0\% accuracy respectively, almost twice of the T2I average of 39\%. T2I models tend to generate more objects than instructed to. \textit{Among} and \textit{Around} naturally accommodate the generation of images featuring multiple objects, providing more flexibility in what can be considered correct. \textit{Beside} also has more flexibility in its answers, and does not have vertical axis constraints (\textit{Above}, \textit{Below}, \textit{On}), depth axis constraints (\textit{In Front Of}, \textit{Behind}), translucency issues (\textit{Inside}) or count constraints (\textit{Between}), allowing for flexibility in placing objects in relation to each other \cite{Retz-Schmidt_1988}.
%, enhancing the model's ability to accurately render scenes. 

% \textbf{SDXL is the worst performing T2I model.} In \Cref{tab:percentage_A_simple}, SDXL only achieves 21.2\% accuracy. In \Cref{fig:simple_results}, it has the most ``D'' ratings (wrong spatial relationship and wrong number/type of objects) and also quite a few ``B'' and ``C'' ratings.

\section{Results: Composing multiple spatial relations}

\begin{figure}[t]
    \centering
    \includegraphics[width=\textwidth]{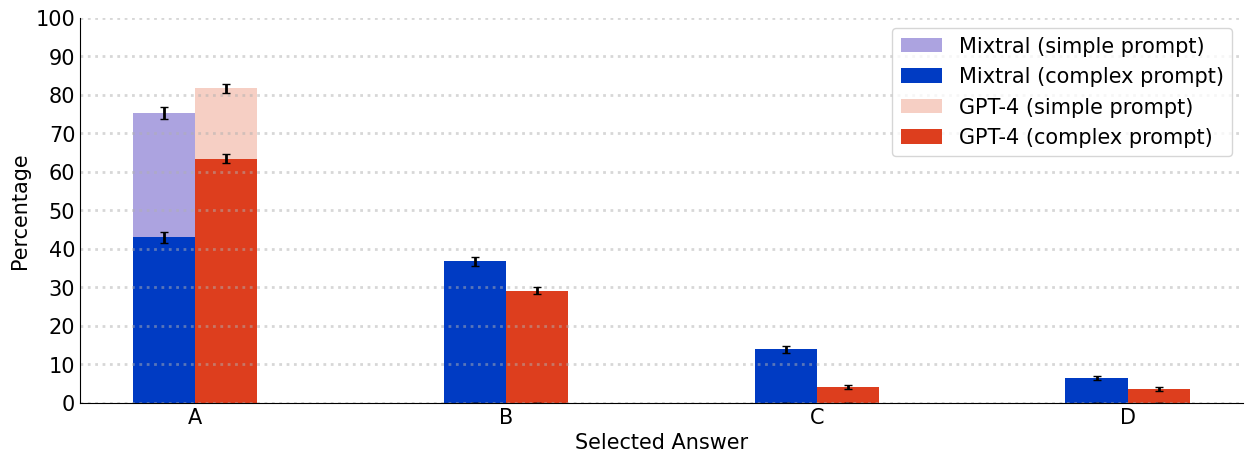}
    \caption{\textbf{Distribution of human ratings for complex prompts.} \textbf{A:} Both spatial relations are correct, and the number and type of objects are all correct. \textbf{B:} Only one spatial relation is correct, but the number and type of objects are all correct. \textbf{C:} Neither of the spatial relations is correct, but the number and type of objects are all correct. \textbf{D:} The number or type of objects are wrong.}
    \label{fig:complex_results}
\end{figure}

We evaluated 2 LLMs on a set of 360 prompts and discuss key findings in \Cref{fig:complex_results} and \Cref{tab:percentage_A_complex}.

% \subsection{Overall}

\begin{table}[ht!]
  \caption{Percentage of ``A'' ratings of images generated for complex prompts.}
  \label{tab:percentage_A_complex}
  \centering
  \fontsize{7.2}{9.5}\selectfont

  \begin{tabularx}{\textwidth}{lccccccccccc}
    \toprule
    \multirow{2}{*}[1.5ex]{\vspace{-8.95ex}Models} & \multicolumn{10}{c}{Spatial Relations} \\
    \cmidrule(lr){2-11}
          & Above & Below & On & Among & Around & Beside & In Front Of & Behind & Between & Inside & \textbf{Average} \\
    \midrule
    Mixtral & 55.6 & 43.6 & 33.8 & 46.3 & 59.8 & 37.2 & \textbf{38.0} & 31.8 & 54.7 & 27.8 & 42.9 \\
    GPT-4 & \textbf{69.7} & \textbf{62.0} & \textbf{47.6} & \textbf{68.9} & \textbf{75.0} & \textbf{64.2} & 37.8 & \textbf{61.5} & \textbf{72.5} & \textbf{74.4} & \textbf{63.4} \\
    \midrule
    \textbf{Average} & 62.6 & 52.8 & 40.7 & 57.6 & 67.4 & 50.7 & 37.9 & 46.6 & 63.6 & 51.1 & 53.1 \\
    \bottomrule
  \end{tabularx}
\end{table}

% \begin{table}[ht!]
%   \caption{Percentage of ``A'' ratings of images generated for complex prompts.}
%   \label{tab:percentage_A_complex}
%   \centering
%   \fontsize{7.2}{9.5}\selectfont

% %   \multicolumn{1}{l}{Parameters} &
% % \multicolumn{2}{c}{Model 1}    &
% % \multicolumn{2}{c}{Model 2}    \\ 
% % \cmidrule(lr){2-3}
% % \cmidrule(lr){4-5}

% % &
% % \multicolumn{1}{c}{Coefficient} &
% % \multicolumn{1}{c}{95\% CI}     &
% % \multicolumn{1}{c}{Coefficient} &
% % \multicolumn{1}{c}{95\% CI}     \\
% % \midrule

%   \begin{tabularx}{\textwidth}{cccccccccccc}
%     \toprule
%      & & \multicolumn{10}{c}{Spatial relations} & 
%      \cmidrule(lr){2-11}
%      Models & Above & Below & On & Among & Around & Beside & In Front Of & Behind & Between & Inside & \textbf{Average}\\
%     \midrule
%     Mixtral & 55.6 & 43.6 & 33.8 & 46.3 & 59.8 & 37.2 & \textbf{38.0} & 31.8 & 54.7 & 27.8 & 42.9 \\
%     GPT-4 & \textbf{69.7} & \textbf{62.0} & \textbf{47.6} & \textbf{68.9} & \textbf{75.0} & \textbf{64.2} & 37.8 & \textbf{61.5} & \textbf{72.5} & \textbf{74.4} & \textbf{63.4} \\
%     \midrule
%     \textbf{Average} & 62.6 & 52.8 & 40.7 & 57.6 & 67.4 & 50.7 & 37.9 & 46.6 & 63.6 & 51.1 & 53.1 \\
%     \bottomrule
%   \end{tabularx}
% \end{table}

\textbf{Performance dropped significantly with complex prompts.} In \Cref{fig:complex_results}, Mixtral had a 32.2\% drop, and GPT-4 had an 18.3\% drop in the percentage of “A” ratings. Compared to Mixtral, GPT-4 is more resilient to extra relations in the complex conditions, maintaining an average accuracy of 63.4\%. This is higher than Llama2, the worst LLM on simple prompts, and DALL·E 3, the best T2I on simple prompts. Therefore, GPT-4 remains reasonably robust to complex prompts.

\textbf{As expected, GPT-4 performed better on complex relations than Mixtral, beating Mixtral on most relations.} This can be seen by the larger proportion of ``A''s attained by GPT-4, averaging 63.4\% as compared to Mixtral with 42.9\% in \Cref{fig:complex_results}. Furthermore, GPT-4 beats Mixtral on 9 out of the 10 spatial relations as can be seen in \Cref{tab:percentage_A_complex}. 

\textbf{Mixtral and GPT-4 struggle with \textit{In Front Of}.} In \Cref{tab:percentage_A_complex}, \textit{In Front Of} scores an average accuracy of 37.9\% compared to the grand average of 53.1\%. This could be due to the ambiguity in perspective as discussed earlier in \Cref{sec:comparison_t2i_llm_results}.

\textbf{Mixtral struggled most with complex prompts containing \textit{Inside}, despite performing well in the simple prompt setting.} In the example ``a cube inside of a cylinder, with a sphere below the cube'', Mixtral makes errors such as making the cube bigger than the cylinder, indicating that it does not understand the conditions and implications of \textit{Inside}, that the cube has to be smaller than the cylinder. This is surprising as Mixtral did well on simple prompts containing \textit{Inside}. For simple prompts containing \textit{Inside}, Mixtral scored 82.5\% accuracy while GPT-4 score 88.8\% (\Cref{tab:percentage_A_simple}).

\textbf{GPT-4 has indicative compositional behaviour as its performance on complex prompts correlates with simple prompts}. While Mixtral's shows insignificant correlation with pearson's coefficient, $r= 0.257$ and a p-value of 0.473. GPT-4 shows fairly strong correlation, with $r=0.706$  and a significant p-value of 0.0224. 

% \begin{figure}[h!]
%     \centering
%     \includegraphics[width=\textwidth]{figures/complex_heatmap_average.jpeg}
%     \caption{Percentage of A rating for complex prompts across each model and spatial relationship.}
%     \label{fig:flat_heatmap}
% \end{figure}

\section{GPT-4V Evaluation}
\begin{figure}[h!]
    \centering
    \includegraphics[width=\textwidth]{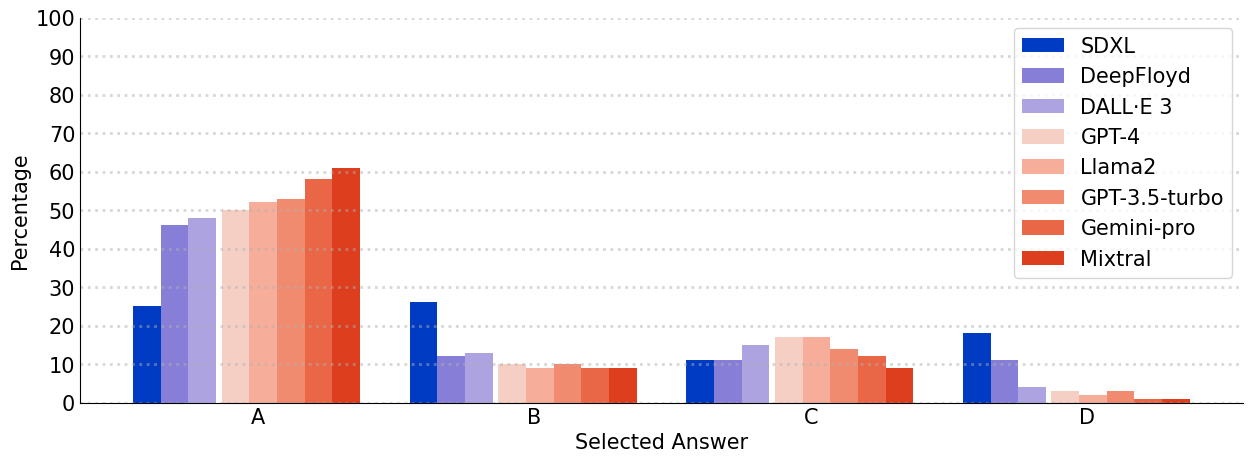}
    % \caption{Auto evaluation results}
    \caption{\textbf{GPT-4V's ratings for simple prompts across each model.} \textbf{A:} Spatial relationship is correct, and the number and type of objects is correct, \textbf{B:} Spatial relationship is correct, but one type of object is wrong, or number of objects is wrong, \textbf{C:} Spatial relationship is wrong, but the type and number of objects are correct, \textbf{D:} Spatial relationship is wrong, and the type of objects are wrong.}
    \label{fig:auto_eval_results}
\end{figure}

The caption setup achieved a Cohen's kappa score of 0.318, while the no caption setup achieved a Cohen's kappa score of 0.326. We demonstrate the overall result of the no caption setup with a higher agreement value in \Cref{fig:auto_eval_results}, while the results for the captioned setup can be found in the Supplementary Material. We found GPT-4V's Cohen kappa score was significantly lower than humans', at 0.326 compared to the human average of 0.483, indicating limitations in GPT-4V's capability as an evaluator. However, GPT-4V is still competitive as compared to open source alternatives, such as Llava, where Llava attained almost no agreement with a cohen's kappa of 0.0140. 

% \section{CLIP evaluation}

% For section 6, around lines 256, can be clearer that GPT-4V being the lowest means the worse compared to leaving out each of the 10 human evaluators. And hence also good to state the lowest human kappa score, i.e. GPT-4V is worse than the worse human?

\section{Conclusion}
% Spatial understanding remains a fundamental aspect of human cognition and evaluating generative models on spatial relations is essential. Evaluations help provide directions to understand and diagnose challenges which current state-of-the-art AI models might face. However, many spatial relations are challenging to express in text alone, making evaluating LLMs on these spatial relations difficult to this day.

We proposed a comprehensive analysis on spatial relations by using a common visual modality to evaluate both LLMs and T2I models. By leveraging on human raters, we were able to cover a wide range of spatial relations and accommodate for multiple relations using complex prompts, providing a deeper understanding
on the spatial relation understanding in generative models.

Our experiments demonstrate that LLMs currently outperform T2I models in spatial relation understanding, with the best performing T2I model performing worse than the worst performing LLM. Performance greatly drops when using complex prompts but GPT-4 is somewhat resilient to handling multiple relations, even showing compositional behaviour by having high correlations between its' performance on complex prompts and simple prompts. Despite performing well in generation, GPT-4 is still weak as an evaluator, having the lowest agreement score when compared to the human evaluators. While recognition is typically easier than generation, there is a difference in the recognition (GPT-4's vision component) and generation component (GPT-4's LLM). This behaviour is different from humans, where the same spatial capabilities are being tested in recognition and in generation.

In summary, we provided a novel perspective in evaluating LLMs and T2I models and demonstrate that there are still large gaps in the understanding of spatial relations in generative models. We believe that the introduction of simple and complex prompts for different levels of relational understanding is a critical avenue for future research and plays an important role in spatial reasoning and intelligence in AI models. Furthermore, it is now possible to evaluate LLMs and T2I in an equal setting, allowing for a more comparative analysis of their capabilities. Lastly, we hope that while automatic evaluation methods are still limited, it will continue to improve and become more relevant in the future.

\bibliographystyle{splncs04}
\bibliography{egbib}

\begin{thebibliography}{10}
\providecommand{\url}[1]{\texttt{#1}}
\providecommand{\urlprefix}{URL }
\providecommand{\doi}[1]{https://doi.org/#1}

\bibitem{achiam2023gpt}
Achiam, J., Adler, S., Agarwal, S., Ahmad, L., Akkaya, I., Aleman, F.L., Almeida, D., Altenschmidt, J., Altman, S., Anadkat, S., et~al.: {GPT-4} technical report. arXiv preprint arXiv:2303.08774  (2023)

\bibitem{agrawal2023stupd}
Agrawal, P., Azaman, H., Tan, C.: Stupd: A synthetic dataset for spatial and temporal relation reasoning. arXiv preprint arXiv:2309.06680  (2023)

\bibitem{ai_2023}
AI: Stability ai (2023), \url{https://stability.ai/news/deepfloyd-if-text-to-image-model}

\bibitem{banerjee-lavie-2005-meteor}
Banerjee, S., Lavie, A.: {METEOR}: An automatic metric for {MT} evaluation with improved correlation with human judgments. In: Proceedings of the {ACL} Workshop on Intrinsic and Extrinsic Evaluation Measures for Machine Translation and/or Summarization. pp. 65--72. Association for Computational Linguistics (2005)

\bibitem{Betker_Goh_Jing}
Betker, J., Goh, G., Jing, L.: \url{https://cdn.openai.com/papers/dall-e-3.pdf}

\bibitem{brown2020language}
Brown, T., Mann, B., Ryder, N., Subbiah, M., Kaplan, J.D., Dhariwal, P., Neelakantan, A., Shyam, P., Sastry, G., Askell, A., et~al.: Language models are few-shot learners. Advances in neural information processing systems  \textbf{33},  1877--1901 (2020)

\bibitem{cho2023dalleval}
Cho, J., Zala, A., Bansal, M.: Dall-eval: Probing the reasoning skills and social biases of text-to-image generation models. In: Proceedings of the IEEE/CVF International Conference on Computer Vision. pp. 3043--3054 (2023)

\bibitem{cobbe2021training}
Cobbe, K., Kosaraju, V., Bavarian, M., Chen, M., Jun, H., Kaiser, L., Plappert, M., Tworek, J., Hilton, J., Nakano, R., et~al.: Training verifiers to solve math word problems. arXiv preprint arXiv:2110.14168  (2021)

\bibitem{comsa-narayanan-2023-benchmark}
Comsa, I., Narayanan, S.: A benchmark for reasoning with spatial prepositions. In: Bouamor, H., Pino, J., Bali, K. (eds.) Proceedings of the 2023 Conference on Empirical Methods in Natural Language Processing. pp. 16328--16335. Association for Computational Linguistics, Singapore (2023)

\bibitem{conwell2022testing}
Conwell, C., Ullman, T.: Testing relational understanding in text-guided image generation. arXiv preprint arXiv:2208.00005  (2022)

\bibitem{coventry1994spatial}
Coventry, K.R., Carmichael, R., Garrod, S.C.: Spatial prepositions, object-specific function, and task requirements. Journal of semantics  \textbf{11}(4),  289--309 (1994)

\bibitem{feng2023layoutgpt}
Feng, W., Zhu, W., Fu, T.J., Jampani, V., Akula, A., He, X., Basu, S., Wang, X.E., Wang, W.Y.: {LayoutGPT}: Compositional visual planning and generation with large language models. In: Advances in Neural Information Processing Systems. vol.~36, pp. 18225--18250. Curran Associates, Inc. (2023)

\bibitem{ghosh2023geneval}
Ghosh, D., Hajishirzi, H., Schmidt, L.: Geneval: An object-focused framework for evaluating text-to-image alignment. In: Advances in Neural Information Processing Systems. vol.~36, pp. 52132--52152. Curran Associates, Inc. (2023)

\bibitem{gisev2013interrater}
Gisev, N., Bell, J.S., Chen, T.F.: Interrater agreement and interrater reliability: key concepts, approaches, and applications. Research in Social and Administrative Pharmacy  \textbf{9}(3),  330--338 (2013)

\bibitem{gokhale2022benchmarking}
Gokhale, T., Palangi, H., Nushi, B., Vineet, V., Horvitz, E., Kamar, E., Baral, C., Yang, Y.: Benchmarking spatial relationships in text-to-image generation. arXiv preprint arXiv:2212.10015  (2022)

\bibitem{hendrycks2020measuring}
Hendrycks, D., Burns, C., Basart, S., Zou, A., Mazeika, M., Song, D., Steinhardt, J.: Measuring massive multitask language understanding. arXiv preprint arXiv:2009.03300  (2020)

\bibitem{Hespos_Spelke_2004}
Hespos, S.J., Spelke, E.S.: Conceptual precursors to language. Nature  \textbf{430}(6998),  453–456 (2004)

\bibitem{hessel2022clipscore}
Hessel, J., Holtzman, A., Forbes, M., Le~Bras, R., Choi, Y.: {CLIPS}core: A reference-free evaluation metric for image captioning. In: Proceedings of the 2021 Conference on Empirical Methods in Natural Language Processing. pp. 7514--7528. Association for Computational Linguistics (2021)

\bibitem{hong2018inferring}
Hong, S., Yang, D., Choi, J., Lee, H.: Inferring semantic layout for hierarchical text-to-image synthesis. In: 2018 IEEE/CVF Conference on Computer Vision and Pattern Recognition (CVPR). pp. 7986--7994 (2018)

\bibitem{huang2023t2i}
Huang, K., Sun, K., Xie, E., Li, Z., Liu, X.: {T2I-CompBench}: A comprehensive benchmark for open-world compositional text-to-image generation. Advances in Neural Information Processing Systems  \textbf{36},  78723--78747 (2023)

\bibitem{jiang2024mixtral}
Jiang, A.Q., Sablayrolles, A., Roux, A., Mensch, A., Savary, B., Bamford, C., Chaplot, D.S., Casas, D.d.l., Hanna, E.B., Bressand, F., et~al.: Mixtral of experts. arXiv preprint arXiv:2401.04088  (2024)

\bibitem{kamath2023s}
Kamath, A., Hessel, J., Chang, K.W.: What's "up" with vision-language models? {I}nvestigating their struggle with spatial reasoning. arXiv preprint arXiv:2310.19785  (2023)

\bibitem{koh2023generating}
Koh, J.Y., Fried, D., Salakhutdinov, R.R.: Generating images with multimodal language models. Advances in Neural Information Processing Systems  \textbf{36} (2023)

\bibitem{tenenbaum_2015}
Lake, B.M., Salakhutdinov, R., Tenenbaum, J.B.: Human-level concept learning through probabilistic program induction. Science  \textbf{350}(6266),  1332–1338 (2015)

\bibitem{lee2023holistic}
Lee, T., Yasunaga, M., Meng, C., Mai, Y., Park, J.S., Gupta, A., Zhang, Y., Narayanan, D., Teufel, H., Bellagente, M., et~al.: Holistic evaluation of text-to-image models. Advances in Neural Information Processing Systems  \textbf{36} (2023)

\bibitem{lian2023llmgrounded}
Lian, L., Li, B., Yala, A., Darrell, T.: {LLM}-grounded diffusion: Enhancing prompt understanding of text-to-image diffusion models with large language models. arXiv preprint arXiv:2305.13655  (2023)

\bibitem{lin-2004-rouge}
Lin, C.Y.: {ROUGE}: A package for automatic evaluation of summaries. In: Text Summarization Branches Out. pp. 74--81. Association for Computational Linguistics, Barcelona, Spain (2004)

\bibitem{lin2024evaluating}
Lin, Z., Pathak, D., Li, B., Li, J., Xia, X., Neubig, G., Zhang, P., Ramanan, D.: Evaluating text-to-visual generation with image-to-text generation. arXiv preprint arXiv:2404.01291  (2024)

\bibitem{TPP}
Litkowski, K., Hargraves, O.: The preposition project. arXiv preprint arXiv:2104.08922  (2021)

\bibitem{liu2022compositional}
Liu, N., Li, S., Du, Y., Torralba, A., Tenenbaum, J.B.: Compositional visual generation with composable diffusion models. In: European Conference on Computer Vision. pp. 423--439. Springer (2022)

\bibitem{lu2024llmscore}
Lu, Y., Yang, X., Li, X., Wang, X.E., Wang, W.Y.: Llmscore: Unveiling the power of large language models in text-to-image synthesis evaluation. Advances in Neural Information Processing Systems  \textbf{36} (2024)

\bibitem{minderer2022simple}
Minderer, M., Gritsenko, A., Stone, A., Neumann, M., Weissenborn, D., Dosovitskiy, A., Mahendran, A., Arnab, A., Dehghani, M., Shen, Z., et~al.: Simple open-vocabulary object detection with vision transformers. arxiv 2022. arXiv preprint arXiv:2205.06230  (2022)

\bibitem{papineni-etal-2002-bleu}
Papineni, K., Roukos, S., Ward, T., Zhu, W.J.: {B}leu: a method for automatic evaluation of machine translation. In: Isabelle, P., Charniak, E., Lin, D. (eds.) Proceedings of the 40th Annual Meeting of the Association for Computational Linguistics. pp. 311--318. Association for Computational Linguistics, Philadelphia, Pennsylvania, USA (2002)

\bibitem{Po2023Compositional3S}
Po, R., Wetzstein, G.: Compositional {3D} scene generation using locally conditioned diffusion. arXiv preprint arXiv:2303.12218  (2023)

\bibitem{podell2023sdxl}
Podell, D., English, Z., Lacey, K., Blattmann, A., Dockhorn, T., M{\"u}ller, J., Penna, J., Rombach, R.: {SDXL}: Improving latent diffusion models for high-resolution image synthesis. arXiv preprint arXiv:2307.01952  (2023)

\bibitem{radford2021learning}
Radford, A., Kim, J.W., Hallacy, C., Ramesh, A., Goh, G., Agarwal, S., Sastry, G., Askell, A., Mishkin, P., Clark, J., et~al.: Learning transferable visual models from natural language supervision. In: International conference on machine learning. pp. 8748--8763. PMLR (2021)

\bibitem{rafailov2024direct}
Rafailov, R., Sharma, A., Mitchell, E., Manning, C.D., Ermon, S., Finn, C.: Direct preference optimization: Your language model is secretly a reward model. Advances in Neural Information Processing Systems  \textbf{36} (2024)

\bibitem{ramesh2022hierarchical}
Ramesh, A., Dhariwal, P., Nichol, A., Chu, C., Chen, M.: Hierarchical text-conditional image generation with clip latents. arXiv preprint arXiv:2204.06125  (2022)

\bibitem{Retz-Schmidt_1988}
Retz-Schmidt, G.: Various views on spatial prepositions. AI Magazine  \textbf{9}(2), ~95 (1988)

\bibitem{rohlfing_nachtigäller_2016}
Rohlfing, K.J., Nachtigäller, K.: Can 28-month-old children learn spatial prepositions robustly from pictures? yes, when narrative input is provided. Frontiers in Psychology  \textbf{7} (2016)

\bibitem{saharia2022photorealistic}
Saharia, C., Chan, W., Saxena, S., Li, L., Whang, J., Denton, E.L., Ghasemipour, K., Gontijo~Lopes, R., Karagol~Ayan, B., Salimans, T., et~al.: Photorealistic text-to-image diffusion models with deep language understanding. Advances in neural information processing systems  \textbf{35},  36479--36494 (2022)

\bibitem{dictionary1989oxford}
Simpson, J.A., C., W.E.S.: The oxford english dictionary. Clarendon (1989)

\bibitem{kuffner2005}
Stilman, M., Kuffner, J.: Navigation among movable obstacles: Real-time reasoning in complex environments. International Journal of Humanoid Robotics  \textbf{02} (2005)

\bibitem{team2023gemini}
Team, G., Anil, R., Borgeaud, S., Wu, Y., Alayrac, J.B., Yu, J., Soricut, R., Schalkwyk, J., Dai, A.M., Hauth, A., et~al.: Gemini: a family of highly capable multimodal models. arXiv preprint arXiv:2312.11805  (2023)

\bibitem{touvron2023llama}
Touvron, H., Martin, L., Stone, K., Albert, P., Almahairi, A., Babaei, Y., Bashlykov, N., Batra, S., Bhargava, P., Bhosale, S., et~al.: Llama 2: Open foundation and fine-tuned chat models. arXiv preprint arXiv:2307.09288  (2023)

\bibitem{turan_mert}
Turan, E., Kobaş, M., Göksun, T.: Spatial language and mental transformation in preschoolers: Does relational reasoning matter? Cognitive Development  \textbf{57},  100980–100980 (2021)

\bibitem{wu2023visual}
Wu, C., Yin, S., Qi, W., Wang, X., Tang, Z., Duan, N.: Visual {ChatGPT}: Talking, drawing and editing with visual foundation models. arXiv preprint arXiv:2303.04671  (2023)

\bibitem{anonymous2023evaluating}
Yamada, Y., Bao, Y., Lampinen, A.K., Kasai, J., Yildirim, I.: Evaluating spatial understanding of large language models. arXiv preprint arXiv:2310.14540  (2023)

\bibitem{Zwarts_2017}
Zwarts, J.: Spatial semantics: Modeling the meaning of prepositions. Language and Linguistics Compass  \textbf{11}(5),  e12241 (2017), e12241 LNCO-0444.R2

\end{thebibliography}

\end{document}